\documentclass[conference]{IEEEtran}
\IEEEoverridecommandlockouts
\usepackage{cite}
\usepackage{amsmath,amssymb,amsfonts}
\usepackage{algorithmic}
\usepackage{graphicx}
\usepackage{textcomp}
\usepackage{xcolor}

\usepackage{times}
\usepackage{algorithm}
\usepackage{multirow}

\def\BibTeX{{\rm B\kern-.05em{\sc i\kern-.025em b}\kern-.08em
    T\kern-.1667em\lower.7ex\hbox{E}\kern-.125emX}}
\begin{document}

\title{NASB: Neural Architecture Search for Binary Convolutional Neural Networks}

\author{\IEEEauthorblockN{Baozhou Zhu$^1$ \hspace{3cm} Zaid Al-Ars$^1$}
\IEEEauthorblockA{
 \\
\hspace*{-3.2cm}\textit{$^1$Delft University of Technology}\\
\hspace*{-3cm}Delft, The Netherlands \\
\hspace*{-3cm}\{b.zhu-1, z.al-ars\}@tudelft.nl}
\and
\IEEEauthorblockN{Peter Hofstee$^{1,2}$}
\IEEEauthorblockA{ \\
\textit{$^2$IBM Research} \\
Austin, TX, USA \\
hofstee@us.ibm.com}
}


\maketitle

\begin{abstract}
Binary Convolutional Neural Networks (CNNs) have significantly reduced the number of arithmetic operations and the size of memory storage needed for CNNs, which makes their deployment on mobile and embedded systems more feasible. However, the CNN architecture after binarizing requires to be redesigned and refined significantly due to two reasons: 1. the large accumulation error of binarization in the forward propagation, and 2. the severe gradient mismatch problem of binarization in the backward propagation. Even though substantial effort has been invested in designing architectures for single and multiple binary CNNs, it is still difficult to find an optimal architecture for binary CNNs. In this paper, we propose a strategy, named NASB, which adopts Neural Architecture Search (NAS) to find an optimal architecture for the binarization of CNNs. Due to the flexibility of this automated strategy,  the obtained architecture is not only suitable for binarization but also has low overhead, achieving a better trade-off between the accuracy and computational complexity of hand-optimized binary CNNs. The implementation of NASB strategy is evaluated on the ImageNet dataset and demonstrated as a better solution compared to existing quantized CNNs. With insignificant overhead increase, NASB outperforms existing single and multiple binary CNNs by up to $4.0\%$ and $1.0\%$ Top-1 accuracy respectively, bringing them closer to the precision of their full precision counterpart. The code and pretrained models will be publicly available.
\end{abstract}

\begin{IEEEkeywords}
binary neural networks, neural architecture search, quantized neural networks, efficiency
\end{IEEEkeywords}

\section{Introduction}
With the increasing depth and width of Convolutional Neural Networks (CNNs), they have demonstrated many breakthroughs in a wide range of applications, such as image classification, object detection, and semantic segmentation \cite{he2016deep,sandler2018mobilenetv2,iandola2016squeezenet}. However, the computation and storage overhead of deep and wide CNNs require millions of FLOPs and parameters, which hinders the real-time deployment on resource-constrained mobile and embedded platforms. 

Numerous researchers proposed different approaches to address the efficiency problem of deploying CNNs, including low bit-width quantization \cite{zhou2016dorefa,zhang2018lq}, network pruning \cite{anwar2017structured}, and efficient architecture design \cite{sandler2018mobilenetv2,zhang2018shufflenet}. Binarization \cite{rastegari2016xnor,liu2018bi} is the most efficient quantization method among all those methods with reduced bit-widths, where a real-valued weight or activation is represented with a single bit and the multiplication and addition of a convolution can be implemented simply by XNOR and popcount bitwise operations, which is roughly $64$ times faster to compute and requires $32$ times less storage than their full precision counterpart. However, the extreme quantization method of single binary CNNs introduces the largest accumulation error in the forward propagation. In addition, during the backward propagation, its gradient flow is the most difficult to determine due to the high gradient mismatch problem~\cite{cai2017deep} among all quantization methods with reduced bit-widths.

Existing published work focuses on improving the quantization quality mainly using value approximation and structure approximation. These two approximations are complementary to each other and could be exploited together. Value approximation seeks to find an optimized algorithm to quantize weights and activations while preserving the original network architecture. Knowledge distillation~\cite{polino2018model,mishra2017apprentice} and loss-aware~\cite{hou2016loss} objectives are introduced to find optimal local minima for quantized weights and activations. Advanced quantization functions~\cite{zhou2016dorefa,cai2017deep,zhang2018lq} are proposed to minimize the quantization error between quantized values and their full precision counterparts. Tight approximation of the derivative of the non-differentiable activation function \cite{liu2018bi,darabi2018bnn} is explored to alleviate the gradient mismatch problem. 

Unlike the above value approximation methods, structure approximation seeks to redesign the architecture of quantized CNNs to match the representational capacity of their original full precision counterpart. Structure approximation is more important for binary CNNs than for other low bit-width CNNs because binarization introduces the largest accumulation error and the severest gradient mismatch problem among all quantization methods with reduced bit-widths. Bi-Real Net~\cite{liu2018bi} and Group-Net~\cite{Zhuang_2019_CVPR} are the state-of-the-art structure approximation methods for single and multiple binary CNNs, respectively. However, designing architectures for quantized CNNs is highly non-trivial especially for binary CNNs. 
In this paper, NASB strategy is proposed to automatically seek an optimal structure approximation for binary CNNs. In particular,  this strategy uses Neural  Architecture  Search  (NAS)  to figure out an optimized architecture for the binarization of CNNs. After searching in a large space, the finalized CNN architecture is suitable for binarization, whose accuracy outperforms previous binary CNNs with insignificant computational complexity increase.

The main contributions in this paper are summarized as follows.
\begin{itemize}
  \item We proposed the NASB strategy, which adopts NAS to automatically find an optimal architecture for the binarization of CNNs. Using NAS, NASB 
  can search in a large space to figure out an optimized CNN architecture, which is suitable for binarization.
  \item Compared to the recent literature of binary CNNs, NASB achieves a sizable accuracy increase with negligible additional overhead, providing a better solution to address the trade-off between accuracy and efficiency.
  \item Our proposed NASB strategy is evaluated for ResNet on the ImageNet classification dataset, providing extensive experimental results to show the effectiveness of our proposal.
\end{itemize}

\section{Related work}
In this section, recent network quantization methods and efficient architecture design developments of CNNs are described.

\subsection{Network quantization}
There is substantial interest in research and development of dedicated hardware for CNNs to be deployed on embedded systems and mobile devices, which motivates the study of network quantization. Low bit-width approaches \cite{zhou2016dorefa,zhang2018lq,faraone2018syq,jung2019learning} quantized weights and activations using fixed-point numbers, which reduces model size and compute time, but still requires multipliers to compute. Binary CNNs \cite{courbariaux2016binarized,rastegari2016xnor,shen2019searching} are trained with weights and activations constrained to binary values $+1$ or $-1$, which can be categorized as single binary CNNs. The Ternary Weight Networks (TWN) \cite{Li2016TernaryWN} approach is proposed to reduce the loss of single binary CNNs by introducing $0$ as the third quantized value, while Trained Ternary Quantization (TTQ) \cite{zhu2016trained} enables the asymmetry and training of its scaling coefficients. However, the accuracy degradation of single binary and ternary CNNs is unacceptable for advanced CNNs like ResNet and large scale datasets like ImageNet. Multiple binary CNNs \cite{fromm2018heterogeneous,lin2017towards,zhu2019binary,Zhuang_2019_CVPR} are promising attempts to reduce the accuracy gap between binary CNNs and their full precision counterpart. However, all the architectures of current single or multiple binary CNNs are human-designed.  Further architecture optimization is possible using automated methods, such as \cite{shen2019searching}, which encodes the number of channels in each layer, but does not change the operations and their connections in the model; something that we do consider in our proposed NASB strategy.

\subsection{Efficient architecture design}
Recently, more and more literature focuses on the efficient architecture design for the deployment of CNNs. Replacing $3 \times 3$ convolutional weights with $1 \times 1$ weights (in SqueezeNet \cite{iandola2016squeezenet} and GoogLeNet \cite{szegedy2015going}) is suggested to decrease the computational complexity. Moreover, separable convolutions are adopted in Inception series \cite{szegedy2017inception} and further generalized as depthwise separable convolutions in Xception \cite{chollet2017xception}, MobileNet \cite{sandler2018mobilenetv2} and ShuffleNet \cite{zhang2018shufflenet}. Group convolution has been used as an efficient way to enhance efficiency in \cite{zhang2018shufflenet,huang2018condensenet},  where the input activations and convolutional kernels are factorized into groups and executed independently inside each group. MobileNet \cite{howard2019searching} and ShuffleNet \cite{ma2018shufflenet} series have been working on depthwise separable convolutions and shuffle operations to achieve a better trade-off between efficiency and accuracy.  ESPNetv2 \cite{mehta2019espnetv2} uses
group point-wise and depth-wise dilated separable convolutions to learn representations from a large effective receptive field, delivering state-of-the-art performance across different tasks. NAS \cite{pham2018efficient,zoph2018learning,cai2018proxylessnas} has demonstrated much success in automating network architecture design, achieving state-of-the-art efficiency \cite{tan2019mnasnet,wu2019fbnet}.

\section{Method}
In this section, the problem of finding an architecture for the binarization of CNNs is defined and presented. Then, we explain NASB strategy, which can adopt the NAS technique to figure out an optimal architecture for binarizing CNNs. Last but not least, variants of NASB strategy are illustrated to enhance its efficiency. 

\subsection{Problem definition}

\begin{figure}[t]
\begin{center}
\includegraphics[width=0.75\linewidth]{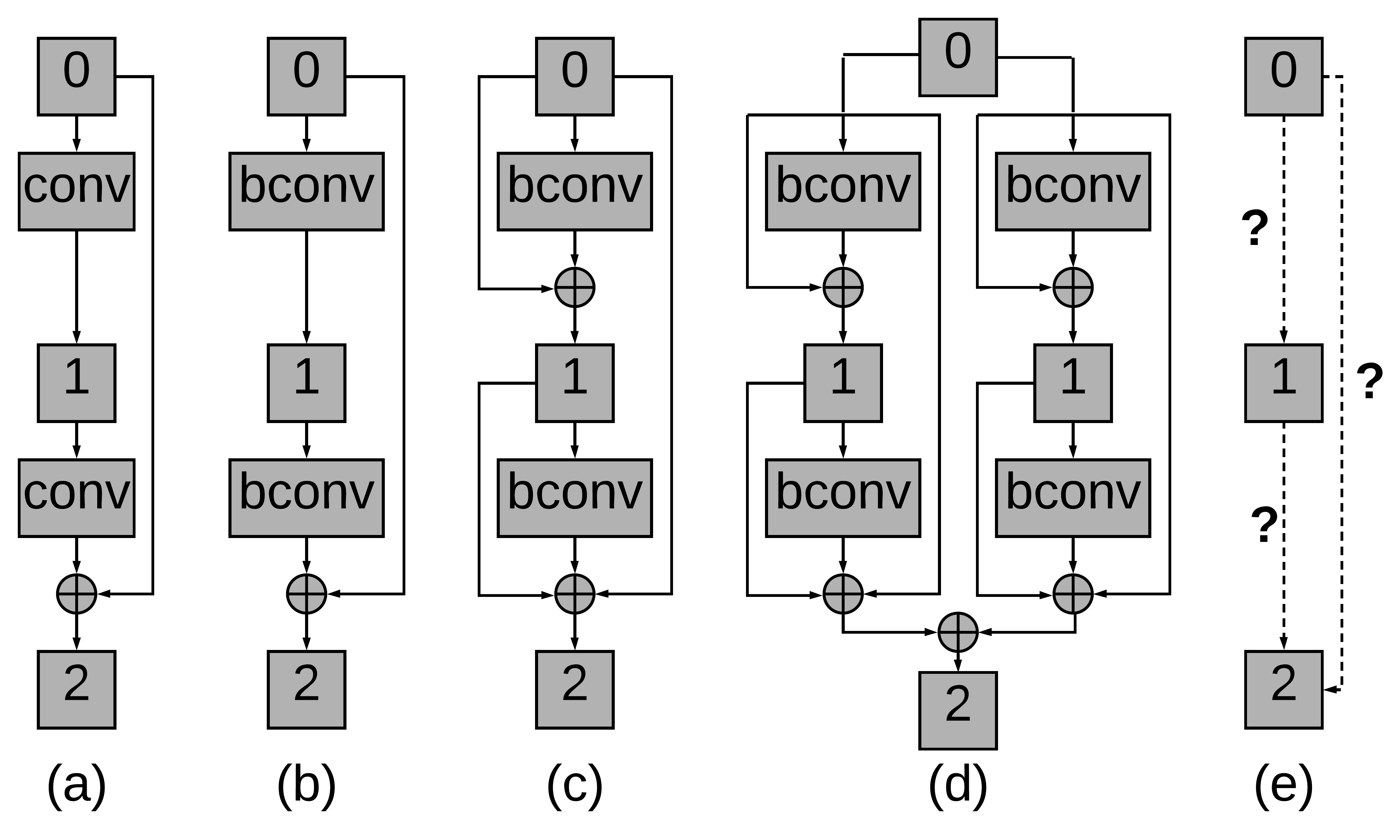}
\end{center}
   \caption{Human-designed architecture for single and multiple binary CNNs. conv and bconv refer to full precision and binary convolutional layer, respectively, while Batch Normalization and the Relu layers are omitted.}
\label{problem}
\end{figure}

Given a full precision convolutional cell, what is an optimal architecture for binarizing it? The accumulation error in the forward propagation of binarization is the largest and the gradient flow in the backward propagation is the most difficult to take care of among all quantization methods with different bit-widths. As a result, it is essential to figure out an optimized architecture for binarizing CNNs. Here, this convolutional cell can be a convolutional layer, block, group, and network.

There have been various attempts to answer the above question, as shown in Fig.~\ref{problem}. Fig.~\ref{problem}(a) is a full precision convolutional block. Fig.~\ref{problem}(b), (c), and (d) describe the proposed architecture in the literature representing XNOR \cite{rastegari2016xnor}, Bi-Real \cite{liu2018bi}, and Group-Net \cite{Zhuang_2019_CVPR}, respectively, where the scaling coefficients have been omitted. Although lots of the above human efforts have been dedicated for designing an architecture for single and multiple binary CNNs, it is still worth to explore an optimal convolutional cell architecture using the automatical approach as represented by the question marks in Fig.~1(e).

The question can be expressed as a directed acyclic graph in Fig.~\ref{problem}(e), which represents an ordered sequence of $3$ nodes and $3$ edges with one operation for each edge.  The number of nodes, edges, and operations for each edge can be freely selected. Each node ${x^i}$ represents a feature map and each edge $(i,j)$ is associated with several operations ${o^{i,j}}$ to transform ${x^i}$. Here the convolutional cell has one input and output node, and its output is obtained by addition of all intermediate nodes. In the following, the binarization and NAS techniques adopted in this paper are presented.

\begin{figure*}[t]
\begin{center}
\includegraphics[width=0.60\linewidth]{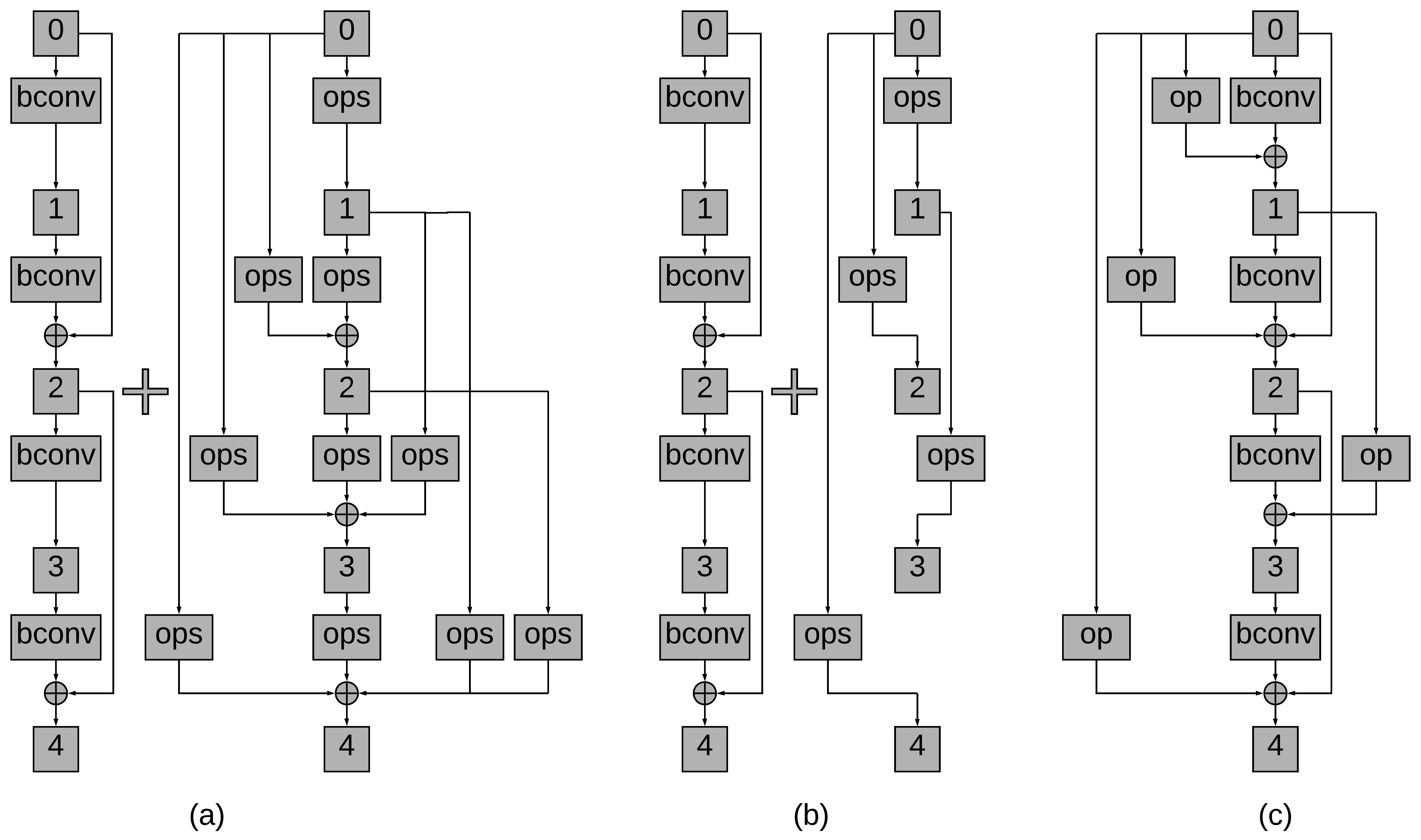}
\end{center}
   \caption{Exploring connections of a NASB-convolutional cell at the searching stage. conv and bconv refer to full precision and binary convolutional layer, respectively. ops refers to a set of operations as shown in Fig.~\ref{ops}, among which one operation is active during the training of the searching stage.
}
\label{NASB}
\end{figure*}

\begin{figure}[t]
\begin{center}
\includegraphics[width=0.70\linewidth]{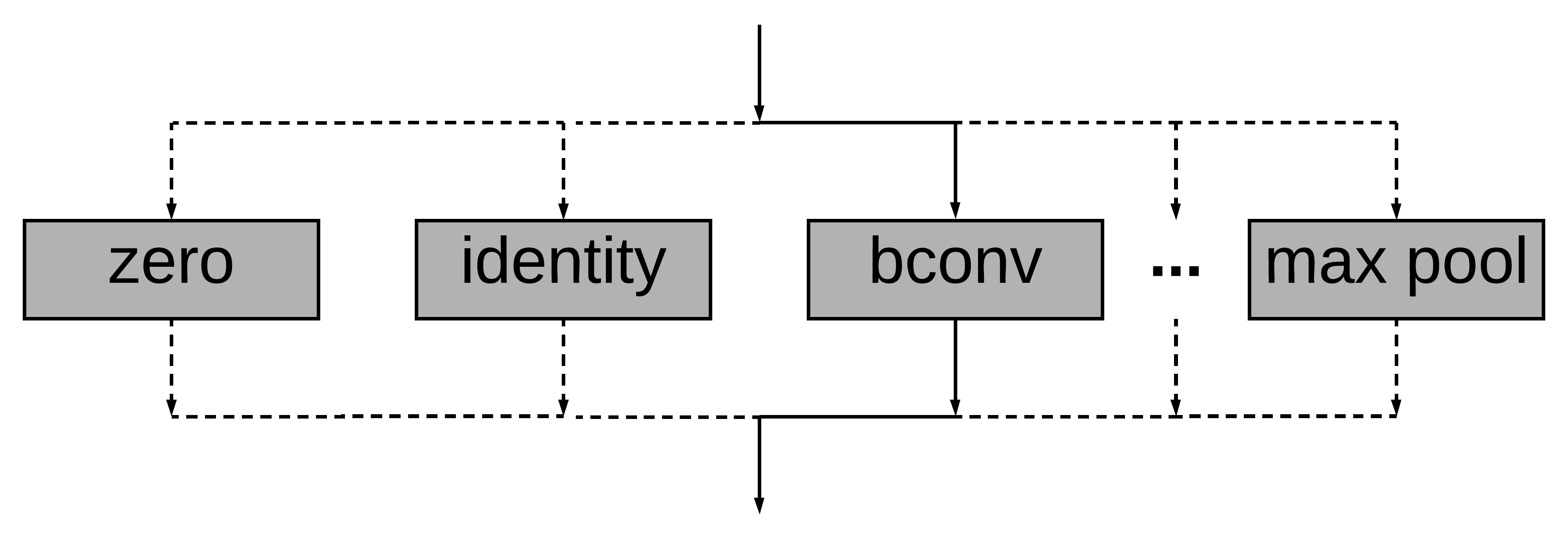}
\end{center}
   \caption{A set of operations in every ops.}
\label{ops}
\end{figure}

\paragraph{Binary convolutional neural networks} \label{Binarization method}
Given a full precision convolutional layer, its inputs, weights and outputs are denoted as $I \in {R^{N \times {C_{in}} \times H \times W}}$, $W \in {R^{{C_{in}} \times {C_{out}} \times h \times w}}$ and $O \in {R^{N \times {C_{out}} \times H \times W}}$, respectively, where $N$, ${C_{in}}$, ${C_{out}}$, $H$, $W$, $h$ and $w$ refer to the batch size, the number of input and output channels, the height and width of the feature maps, and the height and width of the weights, respectively. 

Using the binarization method of weights in \cite{rastegari2016xnor}, we approximate the full precision weights $W$ as binary weights ${b^W}$ with the sign of $W$ and the scaling coefficient $s$, where the scaling coefficient is computed as the mean of the absolute values of $W$. Adopting the Straight Through Estimator (STE) \cite{bengio2013estimating}, the forward and backward propagations of the weights binarization are shown as follows.

\begin{equation}
\begin{split}
\text{Forward: }{b^W} &= s  \times \text{sign}(W)  \\
\text{Backward: }\frac{{\partial L}}{{\partial W}} &= \frac{{\partial L}}{{\partial {b^W}}} \times \frac{{\partial {b^W}}}{{\partial W}} \approx  s \times \frac{{\partial L}}{{\partial {b^W}}} 
\end{split}
\end{equation}
where $L$ is the total loss.

Using the binarization method of activations in \cite{liu2018bi}, we approximate the full precision activations as binary activations ${b^I}$ by a piecewise polynomial function. The forward and backward propagations of the activations binarization can be written as follows.

\begin{equation}
    \begin{split}
\text{Forward: }{b^I} &= \text{sign}(I) \\
\text{Backward: }\frac{{\partial L}}{{\partial I}} &= \frac{{\partial L}}{{\partial {b^I}}} \times \frac{{\partial {b^I}}}{{\partial I}} \\
\text{where}\frac{{\partial {b^I}}}{{\partial I}} &= \left\{ \begin{array}{l}
2 + 2I, - 1 \le I < 0\\
2 - 2I,0 \le I < 1\\
0,\text{otherwise}
\end{array} \right.
    \end{split}
\end{equation}

\paragraph{Gradient based neural architecture search}
We adopt a gradient-based NAS in \cite{liu2018darts}. To reduce memory footprint during training the over-parameterized network, we use the strategy from \cite{cai2018proxylessnas} to binarize and learn the $M$ real-valued architecture parameters ${\alpha _i}$.

In the forward propagation, the $M$ real-valued architecture parameters ${\alpha _i}$ are transformed to the real-valued path weights ${p_i}$, and then to the binary gates ${g_i}$ as follows.

\begin{equation}
{p_i} = \frac{{\exp ({\alpha _i})}}{{\sum\limits_{j = 1}^M {\exp ({\alpha _j})} }}
\end{equation}

\begin{equation}
{g_i} = \text{binarize}({p_i}) = \left\{ \begin{array}{l}
1,\text{with probability }{p_i}\\
0,\text{with probability }(1 - {p_i})
\end{array} \right.
\end{equation}

In the backward propagation, the STE \cite{bengio2013estimating} is also applied.

\begin{equation}
\frac{{\partial L}}{{\partial {p_j}}} \approx \frac{{\partial L}}{{\partial {g_j}}}
\end{equation}

The gradient w.r.t. architecture parameters can be estimated as follows.

\begin{equation}
\begin{split}
\frac{{\partial L}}{{\partial {\alpha _i}}} &= \sum\limits_{j = 1}^M {\frac{{\partial L}}{{\partial {p_j}}}\frac{{\partial {p_j}}}{{\partial {\alpha _i}}}}  \approx \sum\limits_{j = 1}^M {\frac{{\partial L}}{{\partial {g_j}}}\frac{{\partial {p_j}}}{{\partial {\alpha _i}}}} \\
 &= \sum\limits_{j = 1}^M {\frac{{\partial L}}{{\partial {g_j}}}{p_j}({\delta _{ij}} - {p_i})} 
\end{split}
\end{equation}

where ${\delta _{ij}} = 1$ if $i = j$ and ${\delta _{ij}} = 0$ if $i \ne j$.

\subsection{NASB strategy}
In this section, we present the details about how NASB strategy works. To apply NAS for binarizing CNNs, the key innovation is to leverage the NAS technique to find a NASB-convolutional cell as an optimal architecture for binarizing their full precision counterpart, where the NASB-convolutional cell can be a replacement for a binarized convolutional layer, block, group, and network. NASB strategy consists of the following stages: searching stage, pretraining stage, and finetuning stage. In the following, the search space of a NASB-convolutional cell in NASB strategy is described, including its connections and operations. Besides, its training algorithm is presented.

\paragraph{Connections of a NASB-convolutional cell}
Taking that we are exploring an optimal architecture for a convolutional group as an example, the connections of a NASB-convolutional cell in NASB strategy is explored at the searching stage as shown in Fig.~\ref{NASB}. 

Fig.~\ref{NASB}(a) describes all the connections of a NASB-convolutional cell during the training of the searching stage, which consists of a backbone and a NAS-convolutional cell. The left cell is the backbone of the NASB-convolutional cell, which is a standard convolutional group in ResNet \cite{he2016deep}. The right cell is considered as a NAS-convolutional cell, which is a directed acyclic graph consisting of $5$ nodes, $10$ edges, and $10$ operations for every edge. Here $5$ nodes are used to keep the layer depth of a NASB-convolutional cell in NASB strategy the same as its full precision counterpart, which will not increase the latency during inference. The connections of the backbone are fixed and there is no need to specify architecture parameters for it. During the training of the searching stage, the model weights of the NASB-convolutional cell and architecture parameters of the NAS-convolutional cell can be updated alternately, and only one operation on every edge in the NAS-convolutional cell is sampled and active at every step. In this way, the inactive paths reduce the memory requirements. 

Fig.~\ref{NASB}(b) is the finalized architecture after completing the training of the searching stage. In the NAS-convolutional cell, we retain only one predecessor for every node and one operation for every edge except for the node with the number $0$. Fig.~\ref{NASB}(c) is a more compact representation of Fig.~\ref{NASB}(b), showing the output of every node in the NASB-convolutional cell (except for the node with the number $0$) defined as the addition of the two inputs from the backbone and the NAS-convolutional cell.

\paragraph{Operations of a NASB-convolutional cell}
Taking the number of bitwise operations and binary parameters of a $3 \times 3$ binary convolution as $1$ unit, the number of bitwise operations and binary parameters of all the operations used in NASB strategy are unified as shown in Table~\ref{operations}. The overhead of Batch Normalization and Relu layer is not included.

The number of bitwise operations and binary parameters of the binary convolution is $N{C_{out}}HW\times 2{C_{in}}hw$ and ${C_{out}}{C_{in}}hw$, respectively, when no bias is added. Scaling the kernel size of the binary convolution by a scaling coefficient of $s_{k}$, both the number of bitwise operations and binary parameters are scaled by $s_{k}^2$. Changing the dilation rate will not increase the number of bitwise operations and binary parameters of the binary convolution, when the additional cost introduced by padding is omitted. The number of bitwise operations required for computing every individual output of the binary convolution is approximately $2{C_{in}}hw$, while the number of bitwise operations required for computing every individual output of a $3 \times 3$ max and average pooling is $8d$ and $16d$, respectively, where $d$ is the bit-width of pooling operations and $2{C_{in}}hw \gg 16d$ in general. Besides, pooling will not introduce any parameters.

\begin{table}
\caption{The number of bitwise operations and binary parameters of the operations used in NASB. F and B refer to full precision and binary precision, respectively. Bo and Bp refer to Bitwise operations and Binary parameters, respectively.}
\label{operations}
\begin{center}
\begin{tabular}{l|l|l}
\hline
 Operations & Bo & Bp \\
\hline
op0 = Zero (F) & $0$ & $0$\\
op1 = $3 \times 3$ average pooling (F) & $ < 1$  & $0$\\
op2 = $3 \times 3$ max pooling (F) & $ < 1$   & $0$\\
op3  = Identity (F) & $0$ & $0$\\
op4 = $1 \times 1$ convolution (B) & $1/9$  &$1/9$\\
op5 = $3 \times 3$ convolution (B) & $1$   &$1$\\
op6 = $5 \times 5$ convolution (B) &  $25/9$  &  $25/9$\\
op7 = $1 \times 1$ dilated convolution (B) & $1/9$  &$1/9$\\
op8 = $3 \times 3$ dilated convolution (B) & $1$  & $1$\\
op9 = $5 \times 5$ dilated convolution (B) & $25/9$  &$25/9$\\
\hline
\end{tabular}
\end{center}
\end{table}

 \begin{algorithm}[t]
 \caption{Three-stage training algorithm}
 \label{algorithm1}
 \begin{algorithmic}[1]
 \renewcommand{\algorithmicrequire}{\textbf{Input:}}
 \renewcommand{\algorithmicensure}{\textbf{Output:}}
 \REQUIRE Dataset $D$ = $\{ ({X_i},{Y_i})\} _{i = 1}^S$ for the searching stage, dataset $D'$ = $\{ ({X_i'},{Y_i'})\} _{i = 1}^S$ for the pretraining and finetuning stages.
 \ENSURE Binary CNN model ${M_s}$ for the searching stage, full precision CNN model ${M_p}$ for the pretraining stage, and binary CNN model ${M_f}$ for the finetuning stage.
 \\  \textit{} \textbf{Stage 1:} The searching stage
 \FOR {$epoch = 1$ to $L$}
  \FOR {$batch = 1$ to $T$}
   \STATE 
   Randomly sample a mini-batch validation data from $D$, freeze the model weights of model ${M_s}$, and update its architecture parameters. \\ 
   Randomly sample a mini-batch training data from $D$, freeze the architecture parameters of model ${M_s}$, and update its model weights.
  \ENDFOR
  \ENDFOR
\\  \textit{} \textbf{Stage 2:} The pretraining stage
 \FOR {$epoch = 1$ to $L$}
  \FOR {$batch = 1$ to $T$}
   \STATE 
   Randomly sample a mini-batch training data from $D'$  and update the weights of model ${M_p}$.
  \ENDFOR
  \ENDFOR
  \\  \textit{} \textbf{Stage 3:} The finetuning stage
 \FOR {$epoch = 1$ to $L$}
  \FOR {$batch = 1$ to $T$}
   \STATE 
   Randomly sample a mini-batch training data from $D'$ and update the weights of model ${M_f}$.
  \ENDFOR
  \ENDFOR
 \end{algorithmic} 
 \end{algorithm}
\paragraph{Three-stage training algorithm}
As shown in Algorithm~\ref{algorithm1}, the training algorithm of NASB strategy consists of three stages: the searching stage, pretraining stage, and finetuning stage. The goal of the searching stage is to get an optimal binary CNN architecture, which is done by using NAS to train a binary CNN model ${M_s}$ from scratch on dataset $D$. The pretraining stage is used to train a full precision CNN model ${M_p}$ from scratch on dataset $D'$, whose architecture is finalized from the searching stage. The finetuning stage is used to binarize the pre-trained CNN obtained from the pretraining stage and finetune it on dataset $D'$ to get a binary CNN model ${M_f}$. 

The binary CNN model finalized from the searching stage is the same as model ${M_f}$ used in the finetuning stage except for some minor differences because of their different datasets. Performing the searching stage on a small dataset $D$ rather than directly on target dataset $D'$ can be regarded as a proxy task to find the optimal binary architecture model ${M_f}$ for the finetuning stage, which can enable a large search space and significantly accelerate the computation of NASB strategy. After binarizing the full precision CNN model ${M_p}$ from the pretraining stage, we directly get the binary CNN model ${M_f}$ for the finetuning stage.

\subsection{Variants of NASB strategy}
In this section, a number of variants of NASB strategy are presented to improve the accuracy over state-of-the-art multiple binary CNNs. Taking NASB ResNet18 as an example, there are $4$ NASB-convolutional cells, and each of them is composed of $5$ nodes. we retain only one predecessor for every node and one operation for every edge except for the node with the number $0$. By changing the number of NASB-convolutional cells and operations for every node, different variants of NASB strategy are explored.  



\begin{table}[ht]
\begin{center}
\caption{Accuracy of NASB ResNet18 variants}
\label{Variants}
\begin{tabular}{l|l|l}
\hline
Variants & Top-1 & Top-5 \\
\hline 
NASB ResNet18 & $60.5\%$ & $82.2\%$ \\
NASBV1 ResNet18  & $60.3\%$ & $82.3\%$\\
NASBV2 ResNet18 & $61.1\%$ & $82.7\%$\\
NASBV3 ResNet18  & $62.8\%$ & $84.1\%$\\
NASBV4 ResNet18 & $65.3\%$ & $85.9\%$\\
NASBV5 ResNet18 & $66.6\%$ & $87.0\%$\\
\hline
\end{tabular}
\end{center}
\end{table}
\begin{table}[ht]
\caption{Comparisons of ResNet18 with multiple binary methods.}
\label{mulbi}
\begin{center}
\begin{tabular}{l|l|l}
\hline
Model & Top-1 & Top-5 \\
\hline
Full precision  & $69.7\%$ & $89.4\%$\\
ABC-Net ($M = 5$, $N = 5$)  & $65.0\%$ & $85.9\%$\\
Group-Net ($4$ bases)  & $64.2\%$ & $85.6\%$\\
Group-Net** ($4$ bases)  & $66.3\%$ & $86.6\%$\\
NASBV4 & $65.3\%$ & $85.9\%$\\
NASBV5 & $66.6\%$ & $87.0\%$\\
\hline
\end{tabular}
\end{center}
\end{table}
\begin{table}[ht]
\caption{Comparisons with single binary CNNs}
\label{binary}
\begin{center}
\begin{tabular}{l|l|l|l|l|l|l}
\hline
\multicolumn{2}{c|}{Model} & Full & BNN & XNOR & Bi-Real & NASB \\
\hline
\multirow{2}{*}{ResNet18} & Top-1 & $69.7\%$ & $42.2\%$ & $51.2\%$ &  $56.4\%$  &  $60.5\%$ \\
& Top-5 & $89.4\%$ & $67.1\%$ & $73.2\%$ &  $79.5\%$  &  $82.2\%$  \\
\hline
\multirow{2}{*}{ResNet34}  & Top-1 & $73.2\%$ & $-$ & $-$ & $62.2\%$ & $64.0\%$ \\
& Top-5 & $91.4\%$ & $-$ & $-$ & $83.9\%$ & $84.7\%$ \\
\hline
\multirow{2}{*}{ResNet50} & Top-1 & $76.0\%$ & $-$ & $-$ & $62.6\%$ & $65.7\%$ \\
& Top-5 & $92.9\%$ & $-$ & $-$ & $83.9\%$
& $85.8\%$ \\
\hline
\end{tabular}
\end{center}
\end{table}
\begin{table}[ht]
\caption{Comparisons of ResNet18 with fixed-point quantization methods.}
\label{fixed-point}
\begin{center}
\begin{tabular}{l|l|l|l|l}
\hline
Model & W & A & Top-1 & Top-5 \\
\hline
Full precision & $32$ & $32$ & $69.7\%$ & $89.4\%$\\
Dorefa-Net & $2$ & $2$ & $62.6\%$ & $84.4\%$\\
SYQ & $1$ & $8$ & $62.9\%$ & $84.6\%$\\
Lq-Net & $2$ & $2$ & $64.9\%$ & $85.9\%$\\
NASBV4 & $1$  & $1$ & $65.3\%$ & $85.9\%$\\
\hline
\end{tabular}
\end{center}
\end{table}

\begin{figure*}[ht]
\begin{center}
\includegraphics[width=0.70\linewidth]{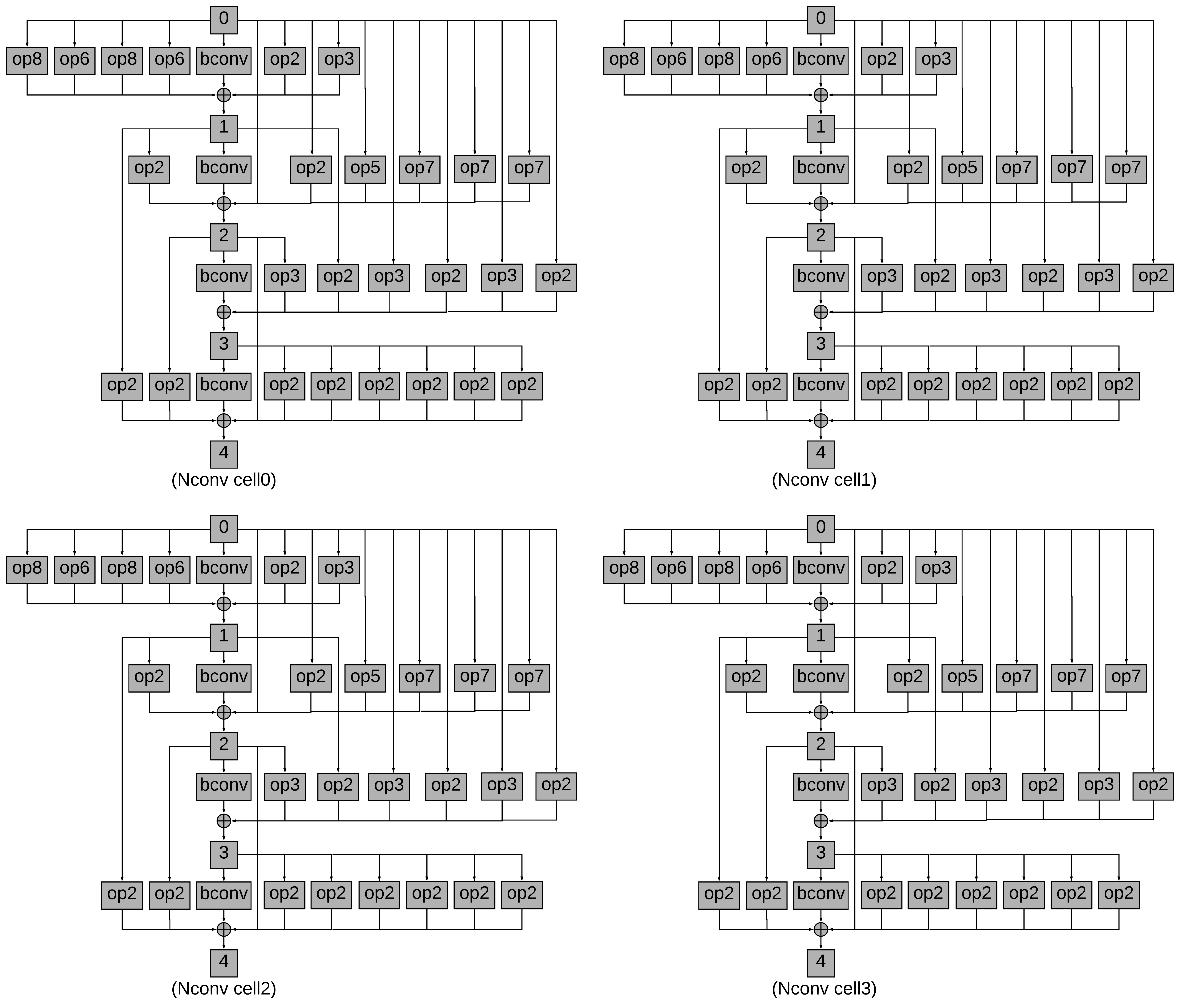}
\end{center}
   \caption{Architecture of NASB-convolutional cells in NASBV5 ResNet18. Nconv cell refers to NASB-convolutional cell. conv and bconv refer to full precision and binary convolutional layer, respectively.}
\label{NASBV5}
\end{figure*}

NASBV1 strategy enlarges the search space of a NASB-convolutional cell. In NASBV1 ResNet18, there are $2$ NASB-convolutional cells, and each of them is composed of $9$ nodes. In NASBV2 ResNet18,  we adopt the method in~\cite{liu2018darts} to retain $4$ operations instead of $1$ operation for the output node of the NASB-convolutional cell. In NASBV3 ResNet18, We copy all the NASB-convolutional cells once to get two binary branches. The two branches can be parallelized thoroughly except that we merge the information of the two branches at the end of every block using addition operation as in \cite{Zhuang_2019_CVPR}. All the NASB-convolutional cells are different from each other, which can explore the optimal binary architecture for every NASB-convolutional cell. In NASBV4 ResNet18, we retain $4$ operations (except for identity) instead of $1$ operation for every node of the NASB-convolutional cell. In NASBV5 ResNet18, we retain $8$ operations for the output node and $6$ operations for the other nodes of the NASB-convolutional cell. Fig.~\ref{NASB}(a) is the connections of a NASB-convolutional cell at the searching stages for the NASB strategy and the NASBV5 strategy. Fig.~~\ref{NASBV5} is the derived architecture of the NASBV5 strategy after the searching stage.




\section{Experimental results on ImageNet dataset}
We applied our proposed NASB strategy for the binarization of ResNet \cite{he2016deep}, trained and evaluated on the ILSVRC2012 classification dataset \cite{russakovsky2015imagenet}. ResNet is one of the most popular and advanced CNNs. 

\subsection{Implementation details}
During the searching stage, we train model  ${M_s}$ on CIFAR-10. Half of the CIFAR-10 training data is used as a validation set. The Relu layer is not added in the searching stage. We train model  ${M_s}$ for $100$ epochs with batch size 64. We use momentum SGD and Adam to optimize the model weights and architecture parameters, respectively. The experiments are performed on one GPU. In NASB ResNet18 and NASB ResNet34,  all NASB-convolutional cells adopt $4$ nodes and they use $3$ nodes for NASB ResNet50. Due to memory limitations, we remove convolutions and dilated convolutions with kernel size $3$ and $5$ for NASB ResNet50 during this stage.

During the pretraining stage, we train model  ${M_p}$ obtained from the last searching stage on the ILSVRC2012 classification dataset. A $224 \times 224$ crop is randomly sampled from an image or its horizontal flip, with the per-pixel mean subtracted. We do not apply any more sophisticated data augmentation to the training data. We use standard single-crop testing for evaluation. We insert the Relu layer and use the layer order as Conv$\to$Relu$\to$BN, and the $tanh$ function is applied to activation after the Batch Normalization layer.

During the finetuning stage, we binarize and train the pre-trained model ${M_p}$ from the pretraining stage into model ${M_f}$. The weights and activations are binarized using the method described in Section~\ref{Binarization method}. We keep $1 \times 1$ convolution to full-precision in this stage. We adopt Adam as the optimizer and set weight decay to $0$ since the binarization can be recognized as a kind of regularization.

\begin{table*}[ht]
\caption{Memory usage and Flops calculation of Bi-Real Net, Group-Net, NASB Net, and full precision models}
\label{table:analysis}
\centering
\begin{tabular}{l|l|l|l|l}
\hline
 Model & Memory usage & Memory saving & Flops & Speedup  \\
\hline
Bi-Real ResNet18  &$33.6$Mbit  &$11.14~\times$  & $ 1.63 \times 10^8
$ & $11.06~\times$ \\
NASB ResNet18 & $33.8$Mbit  & $11.07~\times$ & $1.71 \times 10^8$ & $10.60~\times$ \\
ResNet18  &$ 374.1$Mbit  &$-$  & $ 1.81 \times 10^9
$ & $-$ \\
\hline
Bi-Real ResNet34  &$ 43.7$Mbit  &$15.97~\times$  & $ 1.93 \times 10^8
$ & $18.99~\times$ \\
NASB ResNet34 & $44.0$Mbit  & $15.86~\times$ & $2.01 \times 10^8$ & $18.26~\times$\\
ResNet34  &$ 697.3$Mbit  &$-$  & $3.66 \times 10^9$ & $-$\\
\hline
Bi-Real ResNet50 & $176.8$Mbit  & $4.62~\times$ & $5.45 \times 10^8$ & $7.08~\times$\\
NASB ResNet50 & $178.1$Mbit  & $4.60~\times$ & $6.18 \times 10^8$ & $6.26~\times$\\
ResNet50  &$817.8$Mbit  &$-$  & $3.86 \times 10^9$ & $-$\\
\hline
ABC-Net ($M = 5$, $N = 5$) ResNet18 & $72.3$Mbit  & $5.17~\times$ & $6.74 \times 10^8$ & $2.70~\times$ \\
Group-Net ($4$ bases) ResNet18 & $62.1$Mbit  & $6.03~\times$ & $2.62 \times 10^8$ & $6.90~\times$ \\
Group-Net** ($4$ bases) ResNet18 & $83.9$Mbit  & $4.46~\times$ & $3.38 \times 10^8$ & $5.35~\times$ \\
NASBV4 ResNet18 & $70.7$Mbit  & $5.30~\times$ & $2.81 \times 10^8$ & $6.45~\times$ \\
NASBV5 ResNet18 & $88.3$Mbit  & $4.24~\times$ & $3.52 \times 10^8$ & $5.15~\times$ \\
ResNet18  &$ 374.1$Mbit  &$-$  & $ 1.81 \times 10^9
$ & $-$ \\
\hline
\end{tabular}
\end{table*}

\subsection{Experimental results of NASB variants}
The accuracy of different variants is compared in Table~\ref{Variants}. The accuracy of NASBV1 ResNet18 is almost the same as that of NASB ResNet18. We conjecture that $28/36$ of the total edges in NASBV1 ResNet18 is removed rather than $6/10$ of the total edges in NASB ResNet18, which will change model ${M_s}$ too much and remedy the benefits of a larger search space. For other variants of NASB strategy, we observe the increased operations of NASB-convolutional cell bring Top-1 accuracy improvement by up to $6.0\%$.  It is expected that with more operations retained, NASB variants can achieve higher accuracy. We present the finalized architecture of four NASB-convolutional cells in NASBV5 ResNet18, as shown in Fig.~\ref{NASBV5}, which is derived from Fig.~\ref{NASB}(a) after the searching stage. In this figure, we retain $8$ operations for the output node and $6$ operations for the other nodes of every NASB-convolutional cell. 

\subsection{Comparisons with the state-of-the-art quantized CNNs}
As shown in Table~\ref{binary}, Table~\ref{mulbi}, and Table~\ref{fixed-point}, we compare our NASB strategy with single binary CNNs, multiple parallel binary CNNs, and fixed-point CNNs using different quantization methods, respectively. All the comparison results are directly cited from the corresponding papers. 

As shown in Table~\ref{binary}, Bi-Real Net \cite{liu2018bi} is the state-of-the-art single binary CNNs. Compared with Bi-Real ResNet with varying layers from $18$ to $50$, our proposed NASB ResNet show consistent accuracy improvement by $4.1\%$, $1.8\%$, and $3.1\%$ Top-1 accuracy respectively.

As shown in Table~\ref{mulbi}, we compare our NASB strategy with ABC-Net and Group-Net, which is a multiple binary CNN and can be implemented in a parallel way. Both NASBV4 and NASBV5 achieve higher accuray than ABC-Net. NASBV4 and NASBV5 show better accuracy performance than Group-Net and Group-Net** by $1.1\%$ and $0.3\%$, respectively.

As shown in Table~\ref{fixed-point}, Lq-Net is the current best-performing fixed-point method. Multiple binary CNNs with $K$ binary branches are preferable to fixed-point CNNs with $\sqrt K$ bit-width considering the computational complexity and memory bandwidth \cite{Zhuang_2019_CVPR}. Thus, NASBV4 with $4$ operations retained for every node requires less overhead while still achieves better accuracy.

\subsection{Computational complexity analysis}
To analyze the computational complexity of our proposed NASB strategy, we compare with Bi-Real Net, Group-Net, and full precision models in terms of memory usage saving and computation speedup as shown in Table~\ref{table:analysis}.

The memory usage is computed as the summation of the number of real-valued parameters times $32$ bit and the number of binary parameters times $1$ bit. We use Flops to measure the computation and assume that bitwise XNOR and popcount operations can be calculated in parallel of $64$ on current CPUs. Thus, the Flops is calculated as the summations of $1/64$ of the number of bitwise operations and the number of real-valued operations. Following the suggestion from \cite{rastegari2016xnor,liu2018bi,Zhuang_2019_CVPR}, we keep the first convolutional layer, the last fully connected layer, and the downsampling layer as full precision.

Bi-Real Net \cite{liu2018bi} can be seen as a suboptimal binary CNN architecture of our NASB Net, where one edge connected to its last node is retained for every node and one identity operation remains for every edge. The finalized NAS-convolutional cells in NASB ResNet18 includes $12$ max pooling and $4$ identity operations, and they are composed of $20$ max pooling and $12$ identity operations in NASB ResNet34.  In NASB Res50, the NAS-convolutional cells consist of $41$ max pooling, $6$ identity, and $1$ 1x1 dilated convolution operations. Compared to Bi-Real Net, the increased computational complexity is mainly due to max pooling. The Flops or the number of bitwise operations of a 3x3 max pooling is less than that of a 3x3 convolution, and the additional number of trainable parameters introduced by Batch Normalization of max pooling operation is $2C_{out}$. 

As shown in Table~\ref{table:analysis}, both the additional memory usage and Flops of NASB ResNet of varying depths are negligible compared to Bi-Real Net. ABC-Net requires much more Flops than Group-Net and NASB variants. The increased memory usage and Flops of NASB V5 and NASB V4 ResNet18 are insignificant compared to Group-Net** and Group-Net respectively. 

\section{Conclusion}
In this paper, we proposed a NASB strategy to find an accurate architecture for binary CNNs. Specifically, the NASB strategy uses the NAS technique to identify an optimal architecture in a large search space, which is suitable for binarizing CNNs. We use the ImageNet classification dataset to prove the effectiveness of our proposed approach. With insignificant overhead increases, NASB strategy and its variants achieve up to $4.0\%$ and $1.0\%$ Top-1 accuracy improvement compared with the state-of-the-art single and multiple binary CNNs, respectively, providing a better trade-off between accuracy and efficiency. It is worth to worth to clarify that without we can easily extend our proposed NASB strategy to fixed-point quantized convolutional neural networks and other models for computer vision tasks beyond image classification, which can be explored further in the future. 

\section*{Acknowledgment}

This work was carried out on the Dutch national e-infrastructure with the support of SURF Cooperative.

\bibliographystyle{IEEEtran}
\bibliography{mybibfile}

\begin{thebibliography}{10}
\providecommand{\url}[1]{#1}
\csname url@samestyle\endcsname
\providecommand{\newblock}{\relax}
\providecommand{\bibinfo}[2]{#2}
\providecommand{\BIBentrySTDinterwordspacing}{\spaceskip=0pt\relax}
\providecommand{\BIBentryALTinterwordstretchfactor}{4}
\providecommand{\BIBentryALTinterwordspacing}{\spaceskip=\fontdimen2\font plus
\BIBentryALTinterwordstretchfactor\fontdimen3\font minus
  \fontdimen4\font\relax}
\providecommand{\BIBforeignlanguage}[2]{{%
\expandafter\ifx\csname l@#1\endcsname\relax
\typeout{** WARNING: IEEEtran.bst: No hyphenation pattern has been}%
\typeout{** loaded for the language `#1'. Using the pattern for}%
\typeout{** the default language instead.}%
\else
\language=\csname l@#1\endcsname
\fi
#2}}
\providecommand{\BIBdecl}{\relax}
\BIBdecl

\bibitem{he2016deep}
K.~He, X.~Zhang, S.~Ren, and J.~Sun, ``Deep residual learning for image
  recognition,'' in \emph{Proceedings of the IEEE conference on computer vision
  and pattern recognition}, 2016, pp. 770--778.

\bibitem{sandler2018mobilenetv2}
M.~Sandler, A.~Howard, M.~Zhu, A.~Zhmoginov, and L.-C. Chen, ``Mobilenetv2:
  Inverted residuals and linear bottlenecks,'' in \emph{Proceedings of the IEEE
  Conference on Computer Vision and Pattern Recognition}, 2018, pp. 4510--4520.

\bibitem{iandola2016squeezenet}
F.~N. Iandola, S.~Han, M.~W. Moskewicz, K.~Ashraf, W.~J. Dally, and K.~Keutzer,
  ``Squeezenet: Alexnet-level accuracy with 50x fewer parameters and< 0.5 mb
  model size,'' \emph{arXiv preprint arXiv:1602.07360}, 2016.

\bibitem{zhou2016dorefa}
S.~Zhou, Y.~Wu, Z.~Ni, X.~Zhou, H.~Wen, and Y.~Zou, ``Dorefa-net: Training low
  bitwidth convolutional neural networks with low bitwidth gradients,''
  \emph{arXiv preprint arXiv:1606.06160}, 2016.

\bibitem{zhang2018lq}
D.~Zhang, J.~Yang, D.~Ye, and G.~Hua, ``Lq-nets: Learned quantization for
  highly accurate and compact deep neural networks,'' in \emph{Proceedings of
  the European Conference on Computer Vision (ECCV)}, 2018, pp. 365--382.

\bibitem{anwar2017structured}
S.~Anwar, K.~Hwang, and W.~Sung, ``Structured pruning of deep convolutional
  neural networks,'' \emph{ACM Journal on Emerging Technologies in Computing
  Systems (JETC)}, vol.~13, no.~3, p.~32, 2017.

\bibitem{zhang2018shufflenet}
X.~Zhang, X.~Zhou, M.~Lin, and J.~Sun, ``Shufflenet: An extremely efficient
  convolutional neural network for mobile devices,'' in \emph{Proceedings of
  the IEEE Conference on Computer Vision and Pattern Recognition}, 2018, pp.
  6848--6856.

\bibitem{rastegari2016xnor}
M.~Rastegari, V.~Ordonez, J.~Redmon, and A.~Farhadi, ``Xnor-net: Imagenet
  classification using binary convolutional neural networks,'' in
  \emph{European Conference on Computer Vision}.\hskip 1em plus 0.5em minus
  0.4em\relax Springer, 2016, pp. 525--542.

\bibitem{liu2018bi}
Z.~Liu, W.~Luo, B.~Wu, X.~Yang, W.~Liu, and K.-T. Cheng, ``Bi-real net:
  Binarizing deep network towards real-network performance,'' \emph{arXiv
  preprint arXiv:1811.01335}, 2018.

\bibitem{cai2017deep}
Z.~Cai, X.~He, J.~Sun, and N.~Vasconcelos, ``Deep learning with low precision
  by half-wave gaussian quantization,'' in \emph{Proceedings of the IEEE
  Conference on Computer Vision and Pattern Recognition}, 2017, pp. 5918--5926.

\bibitem{polino2018model}
A.~Polino, R.~Pascanu, and D.~Alistarh, ``Model compression via distillation
  and quantization,'' \emph{arXiv preprint arXiv:1802.05668}, 2018.

\bibitem{mishra2017apprentice}
A.~Mishra and D.~Marr, ``Apprentice: Using knowledge distillation techniques to
  improve low-precision network accuracy,'' \emph{arXiv preprint
  arXiv:1711.05852}, 2017.

\bibitem{hou2016loss}
L.~Hou, Q.~Yao, and J.~T. Kwok, ``Loss-aware binarization of deep networks,''
  \emph{arXiv preprint arXiv:1611.01600}, 2016.

\bibitem{darabi2018bnn}
S.~Darabi, M.~Belbahri, M.~Courbariaux, and V.~P. Nia, ``Bnn+: Improved binary
  network training,'' \emph{arXiv preprint arXiv:1812.11800}, 2018.

\bibitem{Zhuang_2019_CVPR}
B.~Zhuang, C.~Shen, M.~Tan, L.~Liu, and I.~Reid, ``Structured binary neural
  networks for accurate image classification and semantic segmentation,'' in
  \emph{The IEEE Conference on Computer Vision and Pattern Recognition (CVPR)},
  June 2019.

\bibitem{faraone2018syq}
J.~Faraone, N.~Fraser, M.~Blott, and P.~H. Leong, ``Syq: Learning symmetric
  quantization for efficient deep neural networks,'' in \emph{Proceedings of
  the IEEE Conference on Computer Vision and Pattern Recognition}, 2018, pp.
  4300--4309.

\bibitem{jung2019learning}
S.~Jung, C.~Son, S.~Lee, J.~Son, J.-J. Han, Y.~Kwak, S.~J. Hwang, and C.~Choi,
  ``Learning to quantize deep networks by optimizing quantization intervals
  with task loss,'' in \emph{Proceedings of the IEEE Conference on Computer
  Vision and Pattern Recognition}, 2019, pp. 4350--4359.

\bibitem{courbariaux2016binarized}
M.~Courbariaux, I.~Hubara, D.~Soudry, R.~El-Yaniv, and Y.~Bengio, ``Binarized
  neural networks: Training deep neural networks with weights and activations
  constrained to+ 1 or-1,'' \emph{arXiv preprint arXiv:1602.02830}, 2016.

\bibitem{shen2019searching}
M.~Shen, K.~Han, C.~Xu, and Y.~Wang, ``Searching for accurate binary neural
  architectures,'' in \emph{Proceedings of the IEEE International Conference on
  Computer Vision Workshops}, 2019, pp. 0--0.

\bibitem{Li2016TernaryWN}
F.~Li and B.~Liu, ``Ternary weight networks,'' \emph{CoRR}, vol.
  abs/1605.04711, 2016.

\bibitem{zhu2016trained}
C.~Zhu, S.~Han, H.~Mao, and W.~J. Dally, ``Trained ternary quantization,''
  \emph{arXiv preprint arXiv:1612.01064}, 2016.

\bibitem{fromm2018heterogeneous}
J.~Fromm, S.~Patel, and M.~Philipose, ``Heterogeneous bitwidth binarization in
  convolutional neural networks,'' in \emph{Advances in Neural Information
  Processing Systems}, 2018, pp. 4006--4015.

\bibitem{lin2017towards}
X.~Lin, C.~Zhao, and W.~Pan, ``Towards accurate binary convolutional neural
  network,'' in \emph{Advances in Neural Information Processing Systems}, 2017,
  pp. 345--353.

\bibitem{zhu2019binary}
S.~Zhu, X.~Dong, and H.~Su, ``Binary ensemble neural network: More bits per
  network or more networks per bit?'' in \emph{Proceedings of the IEEE
  Conference on Computer Vision and Pattern Recognition}, 2019, pp. 4923--4932.

\bibitem{szegedy2015going}
C.~Szegedy, W.~Liu, Y.~Jia, P.~Sermanet, S.~Reed, D.~Anguelov, D.~Erhan,
  V.~Vanhoucke, and A.~Rabinovich, ``Going deeper with convolutions,'' in
  \emph{Proceedings of the IEEE conference on computer vision and pattern
  recognition}, 2015, pp. 1--9.

\bibitem{szegedy2017inception}
C.~Szegedy, S.~Ioffe, V.~Vanhoucke, and A.~A. Alemi, ``Inception-v4,
  inception-resnet and the impact of residual connections on learning,'' in
  \emph{Thirty-First AAAI Conference on Artificial Intelligence}, 2017.

\bibitem{chollet2017xception}
F.~Chollet, ``Xception: Deep learning with depthwise separable convolutions,''
  in \emph{Proceedings of the IEEE conference on computer vision and pattern
  recognition}, 2017, pp. 1251--1258.

\bibitem{huang2018condensenet}
G.~Huang, S.~Liu, L.~Van~der Maaten, and K.~Q. Weinberger, ``Condensenet: An
  efficient densenet using learned group convolutions,'' in \emph{Proceedings
  of the IEEE Conference on Computer Vision and Pattern Recognition}, 2018, pp.
  2752--2761.

\bibitem{howard2019searching}
A.~Howard, M.~Sandler, G.~Chu, L.-C. Chen, B.~Chen, M.~Tan, W.~Wang, Y.~Zhu,
  R.~Pang, V.~Vasudevan \emph{et~al.}, ``Searching for mobilenetv3,''
  \emph{arXiv preprint arXiv:1905.02244}, 2019.

\bibitem{ma2018shufflenet}
N.~Ma, X.~Zhang, H.-T. Zheng, and J.~Sun, ``Shufflenet v2: Practical guidelines
  for efficient cnn architecture design,'' in \emph{Proceedings of the European
  Conference on Computer Vision (ECCV)}, 2018, pp. 116--131.

\bibitem{mehta2019espnetv2}
S.~Mehta, M.~Rastegari, L.~Shapiro, and H.~Hajishirzi, ``Espnetv2: A
  light-weight, power efficient, and general purpose convolutional neural
  network,'' in \emph{Proceedings of the IEEE Conference on Computer Vision and
  Pattern Recognition}, 2019, pp. 9190--9200.

\bibitem{pham2018efficient}
H.~Pham, M.~Y. Guan, B.~Zoph, Q.~V. Le, and J.~Dean, ``Efficient neural
  architecture search via parameter sharing,'' \emph{arXiv preprint
  arXiv:1802.03268}, 2018.

\bibitem{zoph2018learning}
B.~Zoph, V.~Vasudevan, J.~Shlens, and Q.~V. Le, ``Learning transferable
  architectures for scalable image recognition,'' in \emph{Proceedings of the
  IEEE conference on computer vision and pattern recognition}, 2018, pp.
  8697--8710.

\bibitem{cai2018proxylessnas}
\BIBentryALTinterwordspacing
H.~Cai, L.~Zhu, and S.~Han, ``Proxyless{NAS}: Direct neural architecture search
  on target task and hardware,'' in \emph{International Conference on Learning
  Representations}, 2019. [Online]. Available:
  \url{https://openreview.net/forum?id=HylVB3AqYm}
\BIBentrySTDinterwordspacing

\bibitem{tan2019mnasnet}
M.~Tan, B.~Chen, R.~Pang, V.~Vasudevan, M.~Sandler, A.~Howard, and Q.~V. Le,
  ``Mnasnet: Platform-aware neural architecture search for mobile,'' in
  \emph{Proceedings of the IEEE Conference on Computer Vision and Pattern
  Recognition}, 2019, pp. 2820--2828.

\bibitem{wu2019fbnet}
B.~Wu, X.~Dai, P.~Zhang, Y.~Wang, F.~Sun, Y.~Wu, Y.~Tian, P.~Vajda, Y.~Jia, and
  K.~Keutzer, ``Fbnet: Hardware-aware efficient convnet design via
  differentiable neural architecture search,'' in \emph{Proceedings of the IEEE
  Conference on Computer Vision and Pattern Recognition}, 2019, pp.
  10\,734--10\,742.

\bibitem{bengio2013estimating}
Y.~Bengio, N.~L{\'e}onard, and A.~Courville, ``Estimating or propagating
  gradients through stochastic neurons for conditional computation,''
  \emph{arXiv preprint arXiv:1308.3432}, 2013.

\bibitem{liu2018darts}
H.~Liu, K.~Simonyan, and Y.~Yang, ``Darts: Differentiable architecture
  search,'' \emph{arXiv preprint arXiv:1806.09055}, 2018.

\bibitem{russakovsky2015imagenet}
O.~Russakovsky, J.~Deng, H.~Su, J.~Krause, S.~Satheesh, S.~Ma, Z.~Huang,
  A.~Karpathy, A.~Khosla, M.~Bernstein \emph{et~al.}, ``Imagenet large scale
  visual recognition challenge,'' \emph{International Journal of Computer
  Vision}, vol. 115, no.~3, pp. 211--252, 2015.

\end{thebibliography}
\end{document}